\documentclass[runningheads]{llncs}

\usepackage{accv}

\usepackage{accvabbrv}

\usepackage{graphicx}
\usepackage{booktabs}

\usepackage[accsupp]{axessibility}  

\usepackage[pagebackref,breaklinks,colorlinks,citecolor=accvblue]{hyperref}

\usepackage{orcidlink}
\usepackage{amssymb}
\usepackage{pifont}
\usepackage{makecell}
\usepackage{array}
\usepackage{algorithm}
\usepackage{algorithmicx}%
\usepackage{algpseudocode}%

\definecolor{barrier}{RGB}{112,128,144}
\definecolor{bicycle}{RGB}{220,20,60}
\definecolor{bus}{RGB}{255, 127, 80}
\definecolor{car}{RGB}{255, 158, 0}
\definecolor{const. veh.}{RGB}{233, 150, 70}
\definecolor{motorcycle}{RGB}{255,61,99}
\definecolor{pedestrian}{RGB}{0,0,230}
\definecolor{traffic cone}{RGB}{47,79,79}
\definecolor{trailer}{RGB}{255,140,0}
\definecolor{truck}{RGB}{255,99,71}
\definecolor{drive. suf.}{RGB}{0,207,191}
\definecolor{other flat}{RGB}{175,0,75}
\definecolor{sidewalk}{RGB}{75,0,75}
\definecolor{terrain}{RGB}{112,180,60}
\definecolor{manmade}{RGB}{222,184,135}
\definecolor{vegetation}{RGB}{0,175,0}
\definecolor{others}{RGB}{0, 0, 0}

\begin{document}
\newcolumntype{C}[1]{>{\centering\arraybackslash}m{#1}}
\title{OccFusion: Depth Estimation Free Multi-sensor Fusion for 3D Occupancy Prediction} 

\titlerunning{OccFusion}

\author{Ji Zhang\textsuperscript{$\star$} \and
Yiran Ding\thanks{Equal contribution.} \and
Zixin Liu}

\authorrunning{J.~Zhang et al.}

\institute{\institute{Wuhan University, Hubei, China, 430072 \\
\email{\{jizhang, yrding, liuzixin\}@whu.edu.cn} 
}}

\maketitle

\begin{abstract}
  3D occupancy prediction based on multi-sensor fusion, crucial for a reliable autonomous driving system, enables fine-grained under-
standing of 3D scenes. Previous fusion-based 3D occupancy predictions
relied on depth estimation for processing 2D image features. However,
depth estimation is an ill-posed problem, hindering the accuracy and
robustness of these methods. Furthermore, fine-grained occupancy prediction demands extensive computational resources. To address these issues, we propose OccFusion, a depth estimation free multi-modal fusion framework. Additionally, we introduce a generalizable active training method and an active decoder that can be applied to any occupancy prediction model, with the potential to enhance their performance. Experiments conducted on nuScenes-Occupancy and nuScenes-Occ3D demonstrate our framework's superior performance. Detailed ablation studies highlight the effectiveness of each proposed method.
\end{abstract}

\section{Introduction}
\label{sec:intro}
Accurate and complete perception of 3D surroundings in urban contexts is crucial for autonomous driving, facilitating tasks such as map construction and vehicle motion planning, thereby ensuring safe and reliable driving. Recent years have seen a surge in research on semantic occupancy perception~\cite{iccv01,cvpr01,iccv02,iccv03,3dv01,cvpr02}. Unlike 3D object detection ~\cite{eccv01,eccv02,aaai01,cvpr03}, which typically employs bounding boxes to approximate the location of dynamic objects, semantic occupancy perception models the entire sensor field, encompassing static objects and areas beyond the immediate interest. This approach yields finer-grained 3D scene representations, aligning more closely with real-world driving scenarios, making it a promising research direction.
\begin{figure}[tb]
  \centering
  \includegraphics[height=3.5cm]{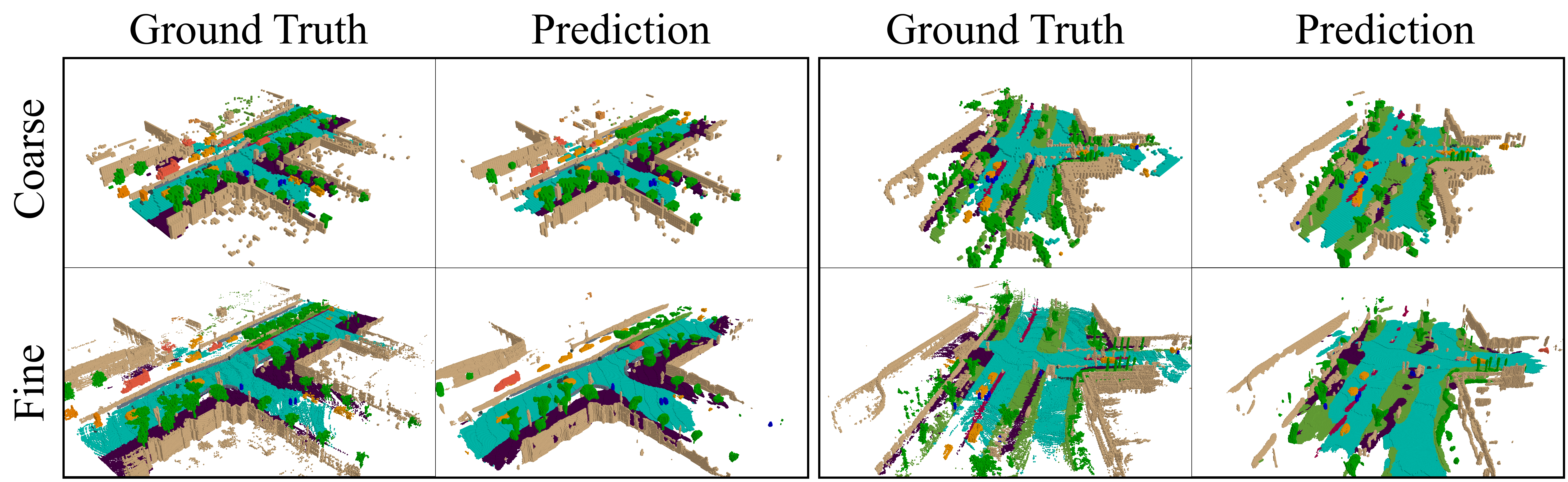}
  \caption{Visualization of our coarse-grained and fine-grained prediction results. The first row shows the ground truth and prediction for two coarse-grained samples, while the second row displays the ground truth and prediction for the same two samples at a fine-grained level. Better viewed when zoomed in.
  }
  \label{fig:coarse fine vis}
\end{figure}

In previous works on semantic surrounding perception ~\cite{iccv02,iccv03,cvpr02,cvpr04,cvpr05,cvpr06,cvpr07,arxiv01,arxiv02,arxiv03,cvpr08,arxiv04}, converting 2D features to 3D through depth prediction has been a conventional approach~\cite{iccv02,cvpr08,arxiv04,cvpr04,cvpr06,cvpr07,arxiv01}. However, it is widely recognized that lifting 2D image features to 3D~\cite{eccv03} inherently attempts to solve an ill-posed problem. The robustness of depth estimation cannot be guaranteed, and considering its use in downstream tasks, the instability of depth estimation poses significant risks to various driving tasks~\cite{eccv01}.

\begin{figure}[tb]
    \centering
    \includegraphics[height=4.5cm]{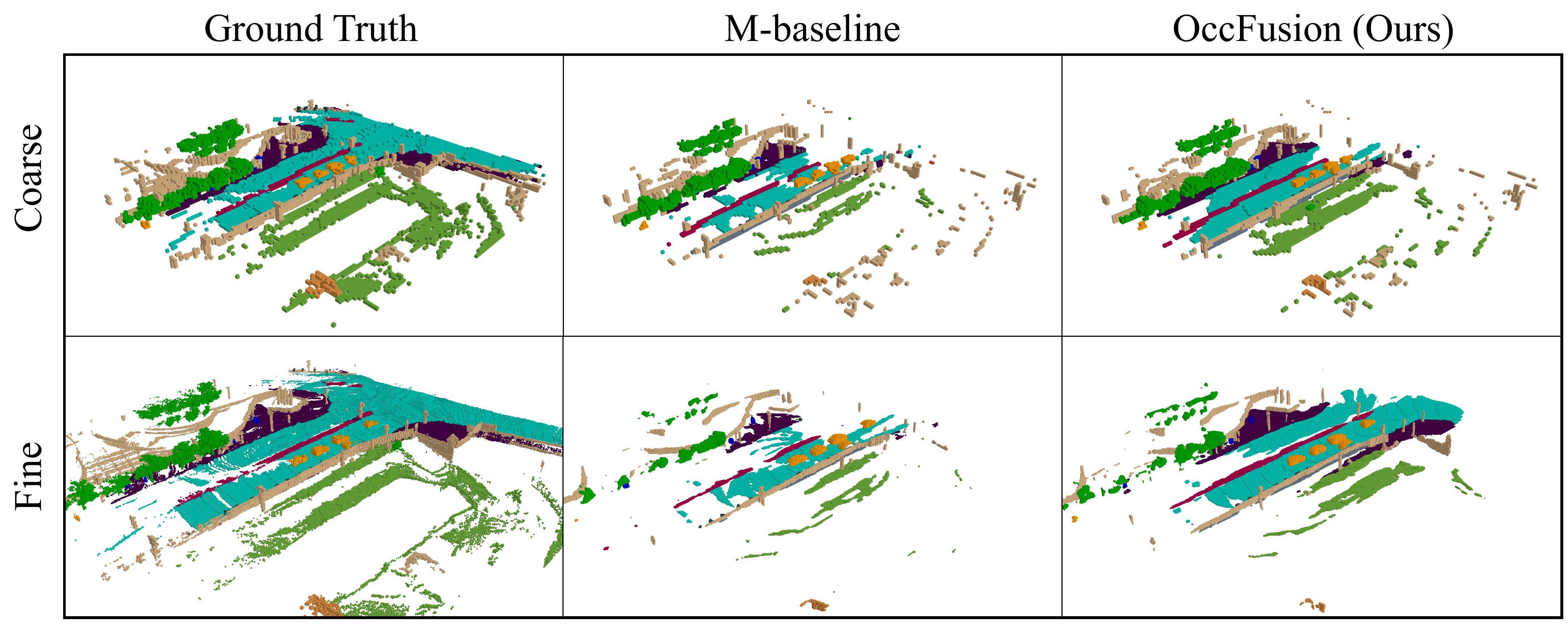}
    \caption{Comparison of our method with one of the existing SOTA multi-modal baseline~\cite{iccv02} under challenging samples. The first row compares M-baseline~\cite{iccv02} with our proposed OccFusion for the coarse occupancy prediction task, while the second row compares M-CONet~\cite{iccv02} with our Active M-CONet. Better viewed when zoomed in.}
    \label{fig:compare}
\end{figure}

By employing multi-modal methods, depth information can be introduced through LiDAR data, mitigating the ill-posed nature of the problem. However, the challenge remains in effectively integrating 2D image features with 3D LiDAR features without depth estimation. While previous literature~\cite{cvpr11,iv01} has indicated that the fusion of multi-modal data can provide redundancy and higher accuracy, to date, only a few studies have focused on multi-modal 3D semantic occupancy prediction~\cite{iccv02,ral01}, and these methods have relied on depth estimation for image features, resulting in suboptimal robustness and accuracy (see \cref{fig:compare}).

On the other hand, some representative fusion-based occupancy prediction methods~\cite{iccv02,ral01} are based on the CONet (\textbf{C}ascade \textbf{O}ccupancy \textbf{Net}work) architecture, which refines all coarse-grained voxels, improving precision while conserving computational resources. However, we point out that splitting operations for most high-confidence voxels are unnecessary and only increase computational load. Additionally, existing models use specific loss functions to address voxel class imbalance from a micro perspective~\cite{iccv02,ral01,neurips04} but overlook the long-tail effect of training data scenes from a macro perspective. New methods are needed to enable the model to selectively learn from more challenging samples, thereby enhancing robustness.

We propose the OccFusion framework, which eliminates depth estimation for image features. Unlike previous methods that blend image features into point cloud features and suffer from density discrepancies between camera and LiDAR features ~\cite{cvpr15,neurips02,cvpr16}, our OccFusion method uses pre-processed LiDAR points to sample image features. Specifically, we voxelize the space around the vehicle and preprocess each voxel's point cloud: for voxels with sparse LiDAR points, we uniformly generate synthetic point clouds; for voxels with dense LiDAR points, we use the farthest point sampling algorithm~\cite{neurips01} to select a subset of points. Then, we project the point cloud onto the image using camera intrinsics and extrinsics to establish correspondences between 2D camera features and 3D LiDAR features. We use respective encoders to obtain 3D and 2D features, and then perform deformable cross attention~\cite{iclr01}, using LiDAR voxel features concatenated with point coordinates as queries and corresponding camera features as keys, directly fusing 3D LiDAR and 2D camera features. The multi-modal features for each LiDAR point within a voxel are averaged to obtain the voxel's multi-modal feature. To achieve fine-grained results, we improve upon CONet~\cite{iccv02} by introducing an active occupancy decoder, which selectively splits challenging voxels to learn fine-grained features, significantly reducing model complexity. Finally, we propose an active training method, allowing the model to prioritize learning from more difficult samples. Experiments show that this simple strategy further improves model performance and can be generalized to the training of other models.

Our contributions can be summarized as follows:
\begin{itemize}
    \item We introduce a novel point-to-point multi-modal feature fusion framework for 3D occupancy prediction, OccFusion, which eliminates the need for depth estimation of image features during the fusion process.
    \item We propose an efficient point cloud preprocessing algorithm that produces denser and more uniform point clouds, which, as demonstrated by experiments, significantly enhances feature fusion.
    \item We propose an active occupancy decoder and an active training method, both of which can be naturally transferred to other occupancy models to improve their performance.
    \item Experiments on nuScenes show that our method achieves state-of-the-art (or comparable) performance across all categories, with significant improvements in the accuracy of small objects. Detailed ablation studies validate the effectiveness of each proposed module and method.
\end{itemize}

\section{Related Work}

\subsection{Vision-Based 3D Occupancy Prediction}
Effectively representing the 3D environment around the vehicle remains a core issue in autonomous driving. Voxel-based representation discretizes 3D space into a voxel grid, computing features for each grid to represent the scene. This method provides finer granularity features than BEV (Bird's Eye View) based methods~\cite{eccv01,eccv02,eccv03,cvpr09,cvpr10}, aligning more closely with real-world driving scenarios. The lack of direct geometric inputs and localization information~\cite{cvpr11}  makes purely camera-based 3D occupancy prediction~\cite{cvpr08, cvpr02} challenging. Recent works~\cite{iccv02,cvpr07,cvpr08,arxiv04,iccv04,arxiv05} have utilized depth prediction methods to generate occupancy features. However, depth prediction is notoriously ill-posed, and these methods often suffer from unstable depth estimation. While camera-only approaches offer promising prospects, multi-modal methods deliver higher accuracy and more reliable results, crucial for the safe and trustworthy deployment of autonomous driving.
\subsection{Feature Fusion of Camera and LiDAR}
LiDAR provides precise localization and reflectance information, complementing camera features~\cite{iccv02,cvpr11,icra01,ral01}. However, LiDAR point clouds are often sparse and vary greatly in density, and lack detailed semantic information such as color and object edges~\cite{eccv02}. Despite the greater expense associated with using multiple sensors, multi-modal semantic occupancy prediction methods~\cite{iccv02,ral01} can integrate the strengths of both LiDAR and camera, outperforming methods based solely on either. However, existing multi-modal approaches face challenges in multi-channel fusion, current works~\cite{iccv02,ral01,icra01} still rely on depth estimation to extract image features, which is an inefficient method. In contrast, our innovative fusion approach does not estimate depth but directly integrates camera and LiDAR features, ensuring efficiency and robustness.

\subsection{Active Learning and Hard Example Mining}
The imbalance of spatial semantic categories and variability in samples can impact model training, hence, random sample selection may degrade accuracy~\cite{ACM01}. Inspired by previous active learning research~\cite{arxiv07,icml01,neurips03}, which suggests that samples with high entropy require additional attention, our active occupancy decoder refines features only for coarse voxels with the highest entropy. During training, we draw inspiration from hard example mining~\cite{cvpr20,tpami01,phd01}, prioritizing samples with greater uncertainty. Experiments demonstrate that this approach significantly improves model accuracy.

\begin{figure}[tb]
    \centering
    \includegraphics[height=6.5cm]{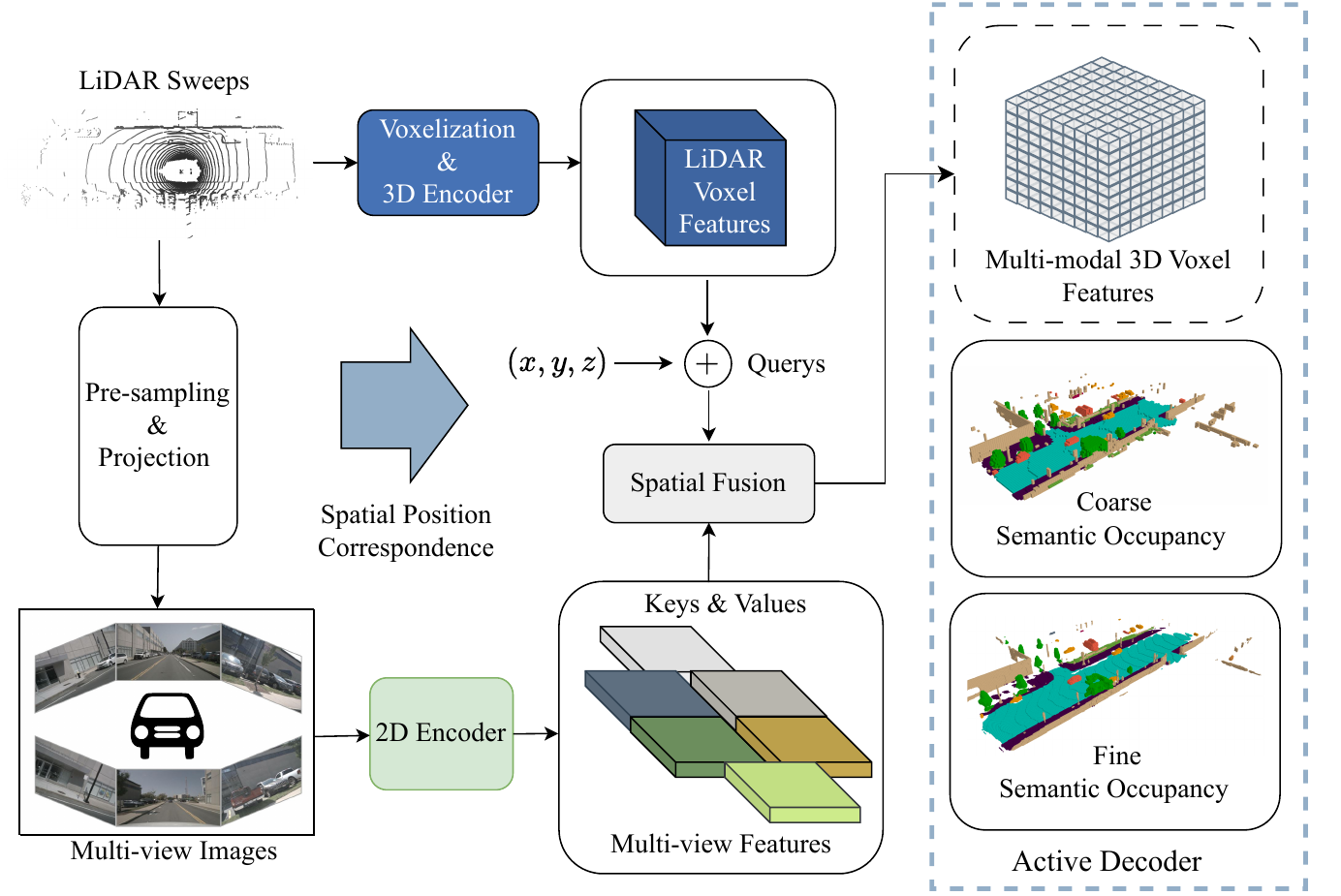}
    \caption{The overall architecture of our method. Raw LiDAR points are processed by a 3D encoder to extract voxelized features, which, concatenated with point coordinates, serve as queries. Multi-view image features, obtained directly through a 2D encoder from surround-view images, act as keys. Enhanced point clouds are then subjected to point-to-point fusion, resulting in multi-modal 3D voxel features. An active decoder adaptively refines predictions in challenging areas.
  }
    \label{fig:architecture}
\end{figure}


\section{Method}
\subsection{Overview}
\Cref{fig:architecture} illustrates the architecture of our method. We employ VoxelNet~\cite{cvpr12} and 3D sparse convolutions~\cite{sensor01} to embed raw LiDAR points into voxelized features $F^L \in R^{\frac{D}{S}\times \frac{H}{S} \times \frac{W}{S} \times C}$(where $S$ is the stride). For camera images, we use ResNet101~\cite{cvpr13} as the backbone to extract multi-view features $F^{mv} \in R^{N \times H^c \times W^c \times C}$, without performing any depth-related operations. With the voxel grid established, the initial state's point cloud is considered to reside within these voxels. Due to the sparsity of raw point clouds, to effectively sample image features, we employ specific sampling and generating methods (see \cref{sec:lidar point sampling}) to ensure each voxel contains dense and relatively uniform LiDAR points. These LiDAR points are then projected onto images using camera intrinsic and extrinsic parameters, creating reference points. With LiDAR points as mediators, we establish the correspondence between LiDAR features and camera features. Through spatial fusion (see \cref{sec:OccFusion}), we fuse 3D LiDAR features concatenated with point coordinates (as queries) with 2D image features (as keys) on a point-to-point basis. For each LiDAR point, we obtain a feature, and after averaging within each voxel, we derive a 
$C$ dimensional feature per voxel. These features can directly yield coarse occupancy predictions through a simple classification head. Due to the high computational complexity of directly predicting refined occupancy grids, we apply an active coarse to fine pipeline to our obtained coarse-grained multi-modal occupancy features, focusing fine-grained prediction only on voxels with the highest uncertainty (see \cref{sec:coarse to fine pipeline}). During the training phase, we experiment with enabling the model to actively learn from samples (see \cref{sec:active training}), which, as experiment shows (see \cref{sec:ablation}), can further improve model accuracy.

\begin{algorithm}
\caption{LiDAR Points Sampling Algorithm}\label{alg:lidar_voxels}

\begin{algorithmic}[1]
\Statex \textbf{Input} Raw LiDAR Point Clouds
\Statex \textbf{Output} 3D Reference Points
\Statex \textbf{Require} $N_p^V \geq 0$, $\tau \in \mathbb{N}$, $\theta \in \mathbb{N}$ ($\theta > \tau$)
\For{each voxel $V$}
    \If{$N_p^V \leq \tau$}
        \State Uniformly generate up to $\theta$ points
    \ElsIf{$\tau < N_p^V \leq \theta$}
        \State \textbf{continue} 
    \Else
        \State Initialize $S \gets \{P_0\}$ with $P_0$ randomly chosen
        \Repeat
            \State $P_j \gets \arg\max\limits_{P \notin S} d(P, S)$
            \State $S \gets S \cup \{P_j\}$
        \Until{$|S| = \theta$}
    \EndIf
\EndFor
\end{algorithmic}
\end{algorithm}

\subsection{3D LiDAR Feature Extraction and LiDAR Point Sampling Algorithm}
\label{sec:lidar point sampling}
The method for embedding raw LiDAR points into 3D voxelized features is consistent with~\cite{iccv02}. In this process, 3D space is partitioned into a grid of size $D/S \times H/S \times W/S$ (where \(S\) is the stride). After partitioning the space, the number of LiDAR points in voxel \(V\) is denoted as $N_p^V$. We define two hyperparameters: $\tau \in \mathbb{N}$ and $\theta \in \mathbb{N}$ ($\theta > \tau$). For each voxel, there are three possible scenarios. First, due to the sparsity of LiDAR point clouds, many voxels contain no or few LiDAR points (\ie, $N_p^V \leq \tau$); for these, we generate synthetic point clouds using a simple uniform generation method to increase the point count to $\theta$. For voxels with an adequate number of LiDAR points (\ie, $\tau < N_p^V \leq \theta$), no action is taken. The uneven spatial distribution of LiDAR point clouds results in some voxels containing too many LiDAR points (\ie, $N_p^V > \theta $); for these, we use farthest point sampling (FPS)~\cite{neurips01} to select $\theta$ points. Specifically, we start with a randomly chosen point $P_0$ as the initial point, forming a sample set $S = \{P_0\}$. We define the distance from a point $P$ to the set as $d(P, S) := \min d(P, P_i), P_i \in S$, calculate the distance $d(P_j, S)$ for all points other than $P_0$, find 
\begin{equation}
    \alpha = \mathop{argmax}\limits_{j}{d(P_j, S), P_j \notin S},
\end{equation} and add $P_\alpha$ to the set $S$, repeating this process until $\theta$ points are obtained. This point cloud sampling algorithm yields denser and more uniformly distributed point clouds in each voxel, facilitating effective sampling of image features.


\subsection{Camera Feature Extraction and OccFusion: Point-to-Point Multi-modal Feature Fusion}

\label{sec:OccFusion}
\begin{figure}[tb]
    \centering
    \includegraphics[height=6.5cm]{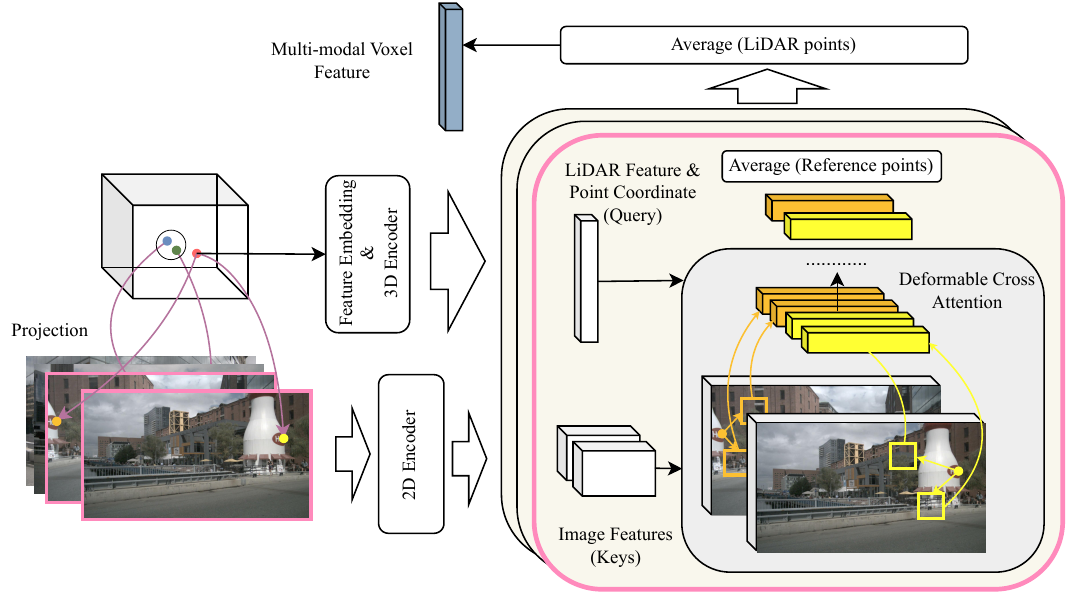}
    \caption{Details of the OccFusion module. After pre-sampling, 3D reference points are projected onto images as (2D) reference points. Note that synthetic point clouds (\emph{points within circles}) do not contribute to LiDAR feature generation. Due to overlapping fields of view among cameras, a single 3D reference point may correspond to multiple reference points upon projection. Features corresponding to reference points are averaged to derive a feature for each 3D reference point, which are then averaged to obtain a multi-modal feature for a voxel.}
    \label{fig:OccFusion}
\end{figure}

We utilize ResNet101~\cite{cvpr13} as the 2D encoder to extract multi-view image features. We do not attempt to lift the spatial dimension of image features to 3D. Instead, we employ a novel OccFusion module to directly fuse 2D image features with 3D LiDAR features on a point-to-point basis. Specifically, we first project the pre-processed point clouds (see \cref{sec:lidar point sampling}) onto multi-view images using camera intrinsic and extrinsic parameters, serving as reference points. Then, for each LiDAR point, we concatenate its coordinates with the LiDAR voxel feature of the voxel where the point is located to form the query. Using this query, we sample and fuse the corresponding image features using deformable attention.~\cite{iclr01}. Bilinear interpolation is used to obtain the sampling features at the corresponding positions. The mechanism of deformable attention and the process mentioned can be formalized as follows:

\begin{equation} DeformAttn(z_q,p_q,x)=\sum_{m=1}^{N_{head}}W_m\sum_{k=1}^{N_{key}}A_{mqk}\cdot W_m^{'}x(p_q+\Delta p_{mqk})
\label{eq:deformable attention}
\end{equation}
and
\begin{equation}
    OccFuse(Q,X,V)=\frac{1}{\left|V\right|}\sum_{l \in {V}}\frac{1}{\left| \mathcal{P}_{L \rightarrow I}(l)\right|}\sum_{i \in \mathcal{P}_{L \rightarrow I}(l)} DeformAttn(Q_{V}^l,i,X_{i}).
    \label{eq:OccFusion}
\end{equation}
In this context, $Q$ represents the 3D LiDAR voxel features, and $X$ is the 2D feature map of surround-view images. The result on the left side of \cref{eq:OccFusion} corresponds to the final feature $F^V$ for a given voxel $V$, where $l$ is a 3D reference point within $V$, and $\mathcal{P}_{L \rightarrow I}(\cdot)$ denotes the projection from the LiDAR coordinate system to the image coordinate system. $\mathcal{P}_{L \rightarrow I}(l)$ is the set of reference points corresponding to the LiDAR point $l$ after projection. Notably, due to the shared field of view among cameras, a single LiDAR point may correspond to multiple reference points across several images upon projection (see \cref{fig:OccFusion}). Moreover, a voxel always contains $|V| \in (\tau, \theta]$ reference points after pre-sampling (see \cref{alg:lidar_voxels}),  and $|\mathcal{P}_{L \rightarrow I}(l)|$ represents the number of reference points corresponding to a single LiDAR point $l$. We use averaging to handle these one-to-many relationships, ultimately obtaining a single feature vector $F^V$ for each voxel. $Q_V^l$ is the query corresponding to voxel $V$ and reference point $l$, and \(X_i\) is the feature map of the image containing reference point \(i\). $W_m$ and $W_m^{^\prime}$ are learnable parameters, $A_{mqk}$ denotes the attention weight. In the prediction phase, the derived feature $F^V$ is processed through a classification head, enabling direct coarse semantic occupancy prediction, see \cref{fig:OccFusion}.

\subsection{Active Coarse to Fine Pipeline}
\label{sec:coarse to fine pipeline}
\begin{figure}[tb]
    \centering
    \includegraphics[height=6.5cm]{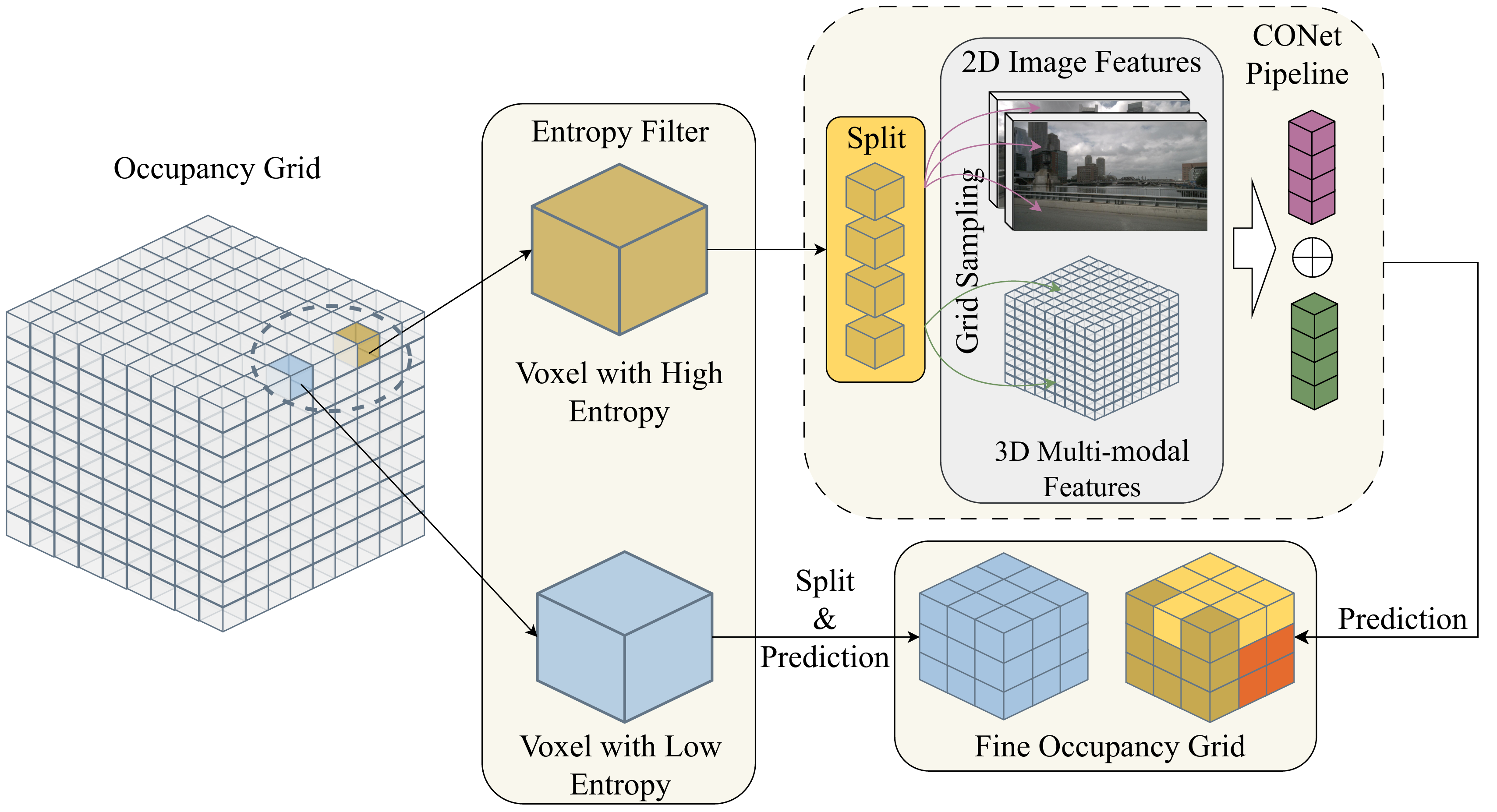}
    \caption{Active coarse to fine pipeline. We refine features only for voxels with greater uncertainty. }
    \label{fig:coarse to fine}
\end{figure}
CONet (Cascade Occupancy Network) saves significant computational resources by refining coarse occupancy instead of directly predicting refined occupancy features~\cite{iccv02,ral01}. However, we note that in real-world scenarios, it is not necessary to refine features for all voxels. For instance, for large objects like buses, a single voxel falling within the space occupied by the bus can sufficiently determine the category of that 3D space with a coarse-grained occupancy feature. Conversely, for very small objects such as traffic cones and bicycles, the coarse-grained occupancy grid is too sparse, making finer-grained prediction crucial. To conserve computational resources while optimizing the model's ability to recognize small or intersecting objects, we introduce an entropy filter to actively determine whether each voxel requires feature refinement. Specifically, we employ classical information entropy to assess the need for fine-grained prediction in a voxel. We set a threshold \(\delta\) representing the proportion of voxels requiring fine-grained feature prediction, and after passing coarse voxel features through a classification head, we obtain probabilities for each class in a voxel $V$ denoted as \(p_i^{V}, i=1,\ldots,N_{class}\), where \(N_{class}\) is the total number of classes. Using the formula
\begin{equation}
    \text{Entropy}(V) = -\sum_{i=1}^{N_{class}}p_i^{V}\log p_{i}^{V},
\end{equation}
we calculate the uncertainty within each coarse-grained voxel. When \(\text{Entropy}(V)\) is sufficiently high, we predict further fine-grained features for the voxel, \ie, splitting the voxel into smaller voxels to serve as occupancy queries for feature sampling through corresponding 2D image features and 3D multi-modal features using camera intrinsic and extrinsic parameters. The details are the same as the coarse to fine pipeline in~\cite{iccv02}. For voxels with lower entropy, we still split them, but for each smaller voxel, we directly use the category obtained from coarse-grained features, see \cref{fig:coarse to fine}. Note that our pipeline can naturally be extended to other occupancy prediction models, not just multi-modal models.

\subsection{Active Training Method}
\label{sec:active training}
\begin{figure}[tb]
    \centering
    \includegraphics[height=6cm]{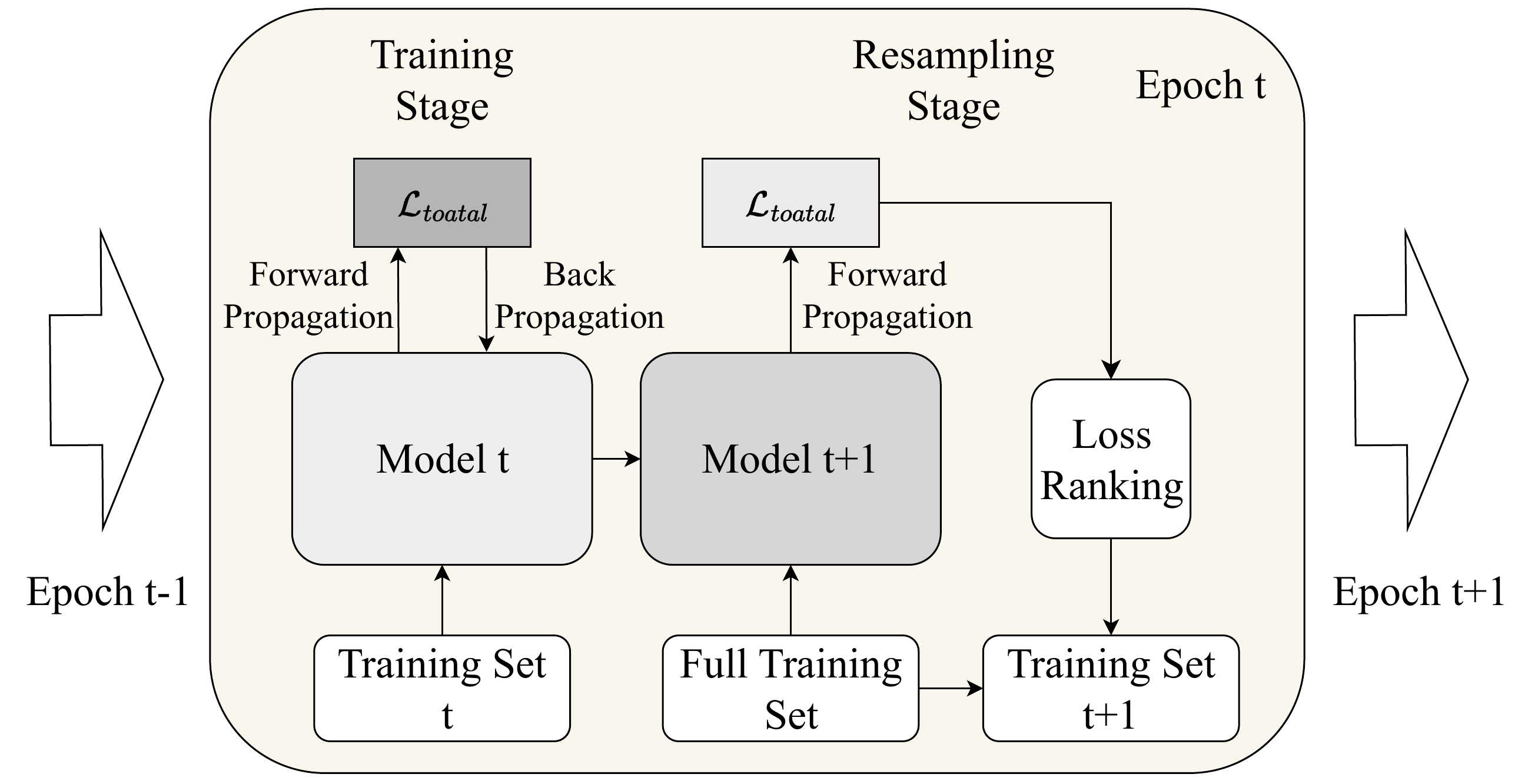}
    \caption{Active training method. In the training stage, we train the model using the training set sampled from the previous stage. In the resampling stage, we use the model trained in the training stage to score the loss on the \emph{full} training set, selecting the top K percent of samples to form the training set for the next training cycle.
}

    \label{fig:active training}
\end{figure}

Due to class imbalance among voxels within each sample, nearly all 3D occupancy prediction methods employ loss functions such as focal loss~\cite{iccv05}, OHEM loss~\cite{cvpr20}, and semantic mIoU~\cite{cvpr08} to enhance model performance on minority semantic classes. However, we note that the complexity of different scenes varies due to factors such as lighting, weather, and surrounding environments. Inspired by classical hard example mining techniques~\cite{cvpr20,tpami01,phd01}, we designed an extremely simple active training approach that biases the model towards difficult samples, which, as demonstrated by experiments, significantly improves model performance. Specifically, at epoch \(t\), starting with the model trained in the previous epoch as model \(t\), training is divided into two stages. In the training stage, we train using the training set \(t\) filtered from the previous epoch to obtain model \(t+1\). Then, using model \(t+1\), we score the loss for each sample in the entire training set and rank these samples from high to low, see \cref{fig:active training}. The higher the loss for a sample,  the greater the necessity for the model to re-learn that sample. Note, in the first round of training, we train the model using all samples. This active training method allows the model to learn from more challenging samples specifically. Although an additional resampling stage for loss ranking is required, this stage does not involve back-propagation, resulting in minimal computational overhead. Moreover, in each epoch (except the first), training only uses the top K percent of samples. Note that our training method is independent of the loss function, meaning it can be combined with any loss function from previous works to enhance model performance.

Here, our model loss is the sum of multiple loss functions, specifically, the cross-entropy loss \(\mathcal{L}_{ce}\), lovasz-softmax \(\mathcal{L}_{ls}\)~\cite{cvpr17}, affinity loss \(\mathcal{L}_{scal}^{geo}\) and \(\mathcal{L}_{scal}^{sem}\)~\cite{cvpr08} (\ie, geometric IoU and semantic mIoU) are combined as the model's loss function, formulated as:
\begin{equation}
    \mathcal{L}_{total} = \mathcal{L}_{ce} + \mathcal{L}_{ls} + \mathcal{L}_{scal}^{geo} + \mathcal{L}_{scal}^{sem}.
    \label{loss}
\end{equation}

\section{Experiments}
\begin{table*}[t]
	\setlength{\tabcolsep}{0.0035\linewidth}
	\newcommand{\classfreq}[1]{{~\tiny(\semkitfreq{#1}\%)}}  %
	\centering
   \caption{Performance on the nuScenes-Occupancy validation set~\cite{iccv02}. \emph{C}, \emph{D}, \emph{L}, \emph{M} represent camera, depth, LiDAR, and multi-modal, respectively. Details of the baseline setup are available in the dataset~\cite{iccv02}.}
   \resizebox{1\linewidth}{!}{
	\begin{tabular}{l| c | c c | c c c c c c c c c c c c c c c c}
 
		\toprule
		Method
		& \makecell[c]{Input}
		& \makecell[c]{IoU}
        & \makecell[c]{mIoU}
		& \rotatebox{90}{\textcolor{barrier}{$\blacksquare$} barrier} 
		& \rotatebox{90}{\textcolor{bicycle}{$\blacksquare$} bicycle}
		& \rotatebox{90}{\textcolor{bus}{$\blacksquare$} bus} 
		& \rotatebox{90}{\textcolor{car}{$\blacksquare$} car} 
		& \rotatebox{90}{\textcolor{const. veh.}{$\blacksquare$} const. veh.} 
		& \rotatebox{90}{\textcolor{motorcycle}{$\blacksquare$} motorcycle} 
		& \rotatebox{90}{\textcolor{pedestrian}{$\blacksquare$} pedestrian} 
		& \rotatebox{90}{\textcolor{traffic cone}{$\blacksquare$} traffic cone} 
		& \rotatebox{90}{\textcolor{trailer}{$\blacksquare$} trailer} 
		& \rotatebox{90}{\textcolor{truck}{$\blacksquare$} truck} 
		& \rotatebox{90}{\textcolor{drive. suf.}{$\blacksquare$} drive. suf.} 
		& \rotatebox{90}{\textcolor{other flat}{$\blacksquare$} other flat} 
		& \rotatebox{90}{\textcolor{sidewalk}{$\blacksquare$} sidewalk} 
		& \rotatebox{90}{\textcolor{terrain}{$\blacksquare$} terrain} 
		& \rotatebox{90}{\textcolor{manmade}{$\blacksquare$} manmade} 
		& \rotatebox{90}{\textcolor{vegetation}{$\blacksquare$} vegetation} \\
		\midrule
		MonoScene~\cite{cvpr08} & C & 18.4 & 6.9 & 7.1  & 3.9  &  9.3 &  7.2 & 5.6  & 3.0  &  5.9& 4.4& 4.9 & 4.2 & 14.9 & 6.3  & 7.9 & 7.4  & 10.0 & 7.6 \\
  
  		TPVFormer~\cite{cvpr02} & C & 15.3 &  7.8 & 9.3  & 4.1  &  11.3 &  10.1 & 5.2  & 4.3  & 5.9 & 5.3&  6.8& 6.5 & 13.6 & 9.0  & 8.3 & 8.0  & 9.2 & 8.2 \\
    
            3DSketch~\cite{cvpr18} &  C\&D & 25.6 & 10.7  & 12.0 &  5.1 &  10.7 &  12.4 & 6.5  & 4.0  & 5.0 & 6.3&  8.0&  7.2& 21.8 &  14.8 & 13.0 &  11.8 & 12.0 & 21.2 \\
            
            AICNet~\cite{cvpr19} & C\&D & 23.8 & 10.6  & 11.5  & 4.0  & 11.8  & 12.3&  5.1 & 3.8  & 6.2  & 6.0 & 8.2&  7.5&  24.1 & 13.0 & 12.8  & 11.5 & 11.6  &  20.2\\

            LMSCNet~\cite{3dv02} & L & 27.3 & 11.5 & 12.4&  4.2 & 12.8  & 12.1  & 6.2  &  4.7 & 6.2 & 6.3&  8.8&  7.2& 24.2 & 12.3  & 16.6 & 14.1  & 13.9 & 22.2 \\

		JS3C-Net~\cite{aaai02} &L & 30.2  & 12.5 & 14.2 & 3.4  & 13.6  & 12.0  & 7.2  &  4.3 & 7.3 & 6.8&  9.2& 9.1 & 27.9 & 15.3  & 14.9 & 16.2  & 14.0 & \textbf{24.9} \\

            C-CONet~\cite{iccv02} & C & 20.1  & 12.8&13.2  & 8.1 &  15.4 &  17.2 & 6.3  & 11.2  & 10.0  &  8.3 & 4.7 & 12.1 & 31.4 & 18.8 & 18.7  & 16.3 & 4.8  &8.2  \\
            
            L-CONet~\cite{iccv02} & L & 30.9  & 15.8 &  17.5  & 5.2  & 13.3  & 18.1  & 7.8  & 5.4  & 9.6 & 5.6& 13.2 & 13.6 & 34.9 & 21.5  & 22.4 & 21.7  & 19.2 &23.5  \\

            M-CONet~\cite{iccv02} & C\&L & 29.5 & 20.1 &  23.3  & 13.3  & 21.2  & 24.3  & 15.3  & 15.9  & 18.0 & 13.3 & 15.3 & 20.7 & 33.2 & 21.0 & 22.5 & 21.5 & 19.6 & 23.2  \\
            
            Co-Occ~\cite{ral01} & C\&L & 30.6 & 21.9 & \textbf{26.5} & 16.8 & 22.3 & \textbf{27.0} & 10.1 & 20.9 & 20.7 & 14.5 & \textbf{16.4} & 21.6 & \textbf{36.9} & \textbf{23.5} & \textbf{25.5} & 23.7 & 20.5 & 23.5 \\
            
            OccFusion (ours) & C\&L  &\textbf{32.4} &\textbf{22.4} & 
25.3 & \textbf{17.0} & \textbf{22.5} & 25.9 & \textbf{16.5} & \textbf{22.4} & \textbf{24.0} & \textbf{16.1} & 16.0 & \textbf{22.1} &
35.6 & 22.1 & 24.0 &
\textbf{23.9} & \textbf{21.3} & 24.0 \tabularnewline
            
		\bottomrule
	\end{tabular}}

	\label{table:base_main}
\end{table*}
\subsection{Experimental Setup}
\subsubsection{Dataset and Metrics}
We conduct experiments on the challenging nuScenes dataset~\cite{cvpr11}, the ground truth labels
are from the works of OpenOccupancy~\cite{iccv02} and Occ3D~\cite{neurips04}. The labels from Occ3D spans a range of $-40 m$ to $40 m$ for the X and Y directions and $-1 m$ to $5.4 m$ for the Z direction, with a voxel size of $0.4m$. For OpenOccupancy, the evaluation range for the X and Y axes is set to $[-51.2m, 51.2m]$, and for the Z axis, it is set to $[-3m, 5m]$. The voxel resolution is 0.2m, resulting in a final occupancy grid spatial scale of $40 \times 512 \times 512$. We utilize the metrics of Intersection-over-Union (IoU) and mean
Intersection-over-Union(mIoU) to evaluate our method's performance. Following ~\cite{iccv02,neurips04}, we train our model on the
training set and evaluated its performance on the validation set.
\begin{table*}[t]
        \setlength{\tabcolsep}{0.0035\linewidth}
	\newcommand{\classfreq}[1]{{~\tiny(\semkitfreq{#1}\%)}}  %
	\centering
\caption{3D semantic occupancy prediction results on Occ3D~\cite{neurips04} benchmark. \emph{C}, \emph{L}, \emph{R} represent camera, LiDAR, and Radar, respectively.}
 
 \resizebox{1\linewidth}{!}{
  \begin{tabular}{c|c|c|ccccccccccccccccc}
    \toprule
    Method & Input & mIoU & 
    \rotatebox{90}{\textcolor{others}{$\bullet$} others} &
    \rotatebox{90}{\textcolor{barrier}{$\bullet$} barrier} & \rotatebox{90}{\textcolor{bicycle}{$\bullet$} bicycle} & \rotatebox{90}{\textcolor{bus}{$\bullet$} bus} & \rotatebox{90}{\textcolor{car}{$\bullet$} car} & \rotatebox{90}{\textcolor{const. veh.}{$\bullet$} const. veh.} & \rotatebox{90}{\textcolor{motorcycle}{$\bullet$} motorcycle} & \rotatebox{90}{\textcolor{pedestrian}{$\bullet$} pedestrian} & \rotatebox{90}{\textcolor{traffic cone}{$\bullet$} traffic cone} & \rotatebox{90}{\textcolor{trailer}{$\bullet$} trailer} & \rotatebox{90}{\textcolor{truck}{$\bullet$} truck} & \rotatebox{90}{\textcolor{drive. suf.}{$\bullet$} drive. surf.} & \rotatebox{90}{\textcolor{other flat}{$\bullet$} other flat} & \rotatebox{90}{\textcolor{sidewalk}{$\bullet$} sidewalk} & \rotatebox{90}{\textcolor{terrain}{$\bullet$} terrain} & \rotatebox{90}{\textcolor{manmade}{$\bullet$} manmade} & \rotatebox{90}{\textcolor{vegetation}{$\bullet$} vegetation} \\
     \midrule 
    MonoScene ~\cite{cvpr08}  & C & 6.06 & 1.75 & 7.23 & 4.26 & 4.93 & 9.38 & 5.67 & 3.98 & 3.01 & 5.90 & 4.45 & 7.17 & 14.91 & 6.32 & 7.92 & 7.43 & 1.01 & 7.65 \\
    BEVDet ~\cite{arxiv06}  & C & 11.73 & 2.09 & 15.29 & 0.0 & 4.18 & 12.97 & 1.35 & 0.0 & 0.43 & 0.13 & 6.59 & 6.66 & 52.72 & 19.04 & 26.45 & 21.78 & 14.51 & 15.26 \\
    BEVFormer ~\cite{eccv01}  & C & 23.67 & 5.03 & 38.79 & 9.98 & 34.41 & 41.09 & 13.24 & 16.50 & 18.15 & 17.83 & 18.66 & 27.70 & 48.95 & 27.73 & 29.08 & 25.38 & 15.41 & 14.46 \\
    BEVStereo ~\cite{aaai03}  & C & 24.51 & 5.73 & 38.41 & 7.88 & 38.70 & 41.20 & 17.56 & 17.33 & 14.69 & 10.31 & 16.84 & 29.62 & 54.08 & 28.92 & 32.68 & 26.54 & 18.74 & 17.49 \\
    TPVFormer ~\cite{cvpr02}  & C & 28.34 & 6.67 & 39.20 & 14.24 & 41.54 & 46.98 & 19.21 & 22.64 & 17.87 & 14.54 & 30.20 & 35.51 & 56.18 & 33.65 & 35.69 & 31.61 & 19.97 & 16.12 \\
    OccFormer ~\cite{iccv04}  & C & 21.93 & 5.94 & 30.29 & 12.32 & 34.40 & 39.17 & 14.44 & 16.45 & 17.22 & 9.27 & 13.90 & 26.36 & 50.99 & 30.96 & 34.66 & 22.73 & 6.76 & 6.97 \\
    CTF-Occ ~\cite{neurips04} & C & 28.53 & 8.09 & 39.33 & 20.56 & 38.29 & 42.24 & 16.93 & 24.52 & 22.72 & 21.05 & 22.98 & 31.11 & 53.33 & 33.84 & 37.98 & 33.23 & 20.79 & 18.00 \\
    RenderOcc ~\cite{arxiv13}  & C & 26.11 & 4.84 & 31.72 & 10.72 & 27.67 & 26.45 & 13.87 & 18.20 & 17.67 & 17.84 & 21.19 & 23.25 & 63.20 & 36.42 & 46.21 & 44.26 & 19.58 & 20.72 \\
    BEVDet4D ~\cite{arxiv12}  & C & 42.02 & 12.15 & 49.63 & 25.10 & 52.02 & 54.46 & 27.87 & 27.99 & 28.94 & 27.23 & 36.43 & 42.22 & 82.31 & 43.29 & 54.46 & 57.90 & 48.61 & 43.55 \\
    PanoOcc ~\cite{arxiv11} & C & 42.13 & 11.67 & 50.48 & 29.64 & 49.44 & 55.52 & 23.29 & 33.26 & 30.55 & 30.99 & 34.43 & 42.57 & 83.31 & 44.23 & 54.40 & 56.04 & 45.94 & 40.40 \\
    FB-OCC ~\cite{arxiv10}  & C & 43.41 & 12.10 & 50.23 & 32.31 & 48.55 & 52.89 & 31.20 & 31.25 & 30.78 & 32.33 & 37.06 & 40.22 & \textbf{83.34} & \textbf{49.27} & \textbf{57.13} & 59.88 & 47.67 & 41.76 \\
    OctreeOcc ~\cite{arxiv09} & C & 44.02 & 11.96 & 51.70 & 29.93 & 53.52 & 56.77 & 30.83 & 33.17 & 30.65 & 29.99 & 37.76 & 43.87 & 83.17 & 44.52 & 55.45 & 58.86 & 49.52 & 46.33 \\
    \hline
    Ming \etal ~\cite{arxiv08} & C+L+R & 46.67 & \textbf{12.37} & 50.33 & 31.53 & \textbf{57.62} & \textbf{58.81} & 33.97 & 41.00 & 47.18 & 29.67 & \textbf{42.03} & 48.04 & 78.39 & 35.68 & 47.26 & 52.74 & \textbf{63.46} & \textbf{63.30} \\
    OccFusion (ours) & C+L & \textbf{48.74} & 12.35 & \textbf{51.77} & \textbf{33.01} & 54.56 & 57.65& \textbf{33.99} & \textbf{43.03} & \textbf{48.35} &\textbf{35.54} & 41.22 & \textbf{48.55}& 83.00 & 44.65 & \textbf{57.13} & \textbf{60.01} & 62.46 & 61.25 \\
    
    \bottomrule
  \end{tabular}}
  \label{table:occ_3d}
\end{table*}

\subsubsection{Implementation Details}
Our method is based on CONet and is highly comparable to M-CONet~\cite{iccv02} and Co-Occ~\cite{ral01}.
For fairness, we adopt a foundational setup largely similar to ~\cite{iccv02}. Specifically, we utilize an ImageNet~\cite{cvpr14} pretrained ResNet101~\cite{cvpr13} with FPN~\cite{cvpr21} as the 2D encoder for images, with an input image size of $1600 \times 900$. For LiDAR branch, we voxelize 10 LiDAR sweeps and use \cite{cvpr12,sensor01} as 3D encoder. During the training phase, we employ the AdamW~\cite{iclr02} optimizer, with weight decay and initial learning rate set to 0.01 and 2e-4, respectively. A cosine learning rate scheduler with linear warm-up in the first 500 iterations is leveraged. Image augmentation strategies follow those used in BEVDet~\cite{arxiv06}. In the point pre-sampling process, hyper-parameters $\theta$ and $\tau$ are set to 20 and 5, respectively. Our model is trained for 24 epochs on 8 A100 GPUs with a batch size of 8. Moreover, the effectiveness of the active training method proposed above (see \cref{sec:active training}) is tested separately (see \cref{tab:my_label03}).

\subsection{3D Semantic Occupancy Prediction}
We compare our method with recent state-of-the-art (SOTA) models, with accuracy data provided by ~\cite{ral01,arxiv08}—all results reported by the authors themselves or obtained using open-source code (see \cref{table:base_main,table:occ_3d}). Across two benchmarks, our method achieves SOTA mIoU and notably increases mIoU by a large margin (2.07) on Occ3D. Our approach also excels in small object categories. Specifically, on the nuscenes-occupancy benchmark, our model's mIoU improves relative to M-CONet~\cite{iccv02} by 27.8\% for bicycle, 40.9\% for motorcycle, 33.3\% for pedestrian, and 21.1\% for traffic cone. On Occ3D, our model also significantly enhances precision in most small object categories. Notably, compared to M-CONet, our model reduces computational load by about 70\% during the coarse to fine phase and maintains complexity comparable to vision-based methods (see \cref{tab:my_label05,tab:my_label04}). On both benchmarks, compared to the best existing uni-modal methods, our model respectively achieves a relative mIoU increase of 41.8\% and 10.7\%, showcasing the significant advantages of multi-modal approaches. 
\subsection{Ablation Study}
\label{sec:ablation}
To validate the effectiveness of the methods we proposed and the computational complexity of our model, we conduct extensive ablation experiments on the OpenOccupancy benchmark~\cite{iccv02}.
\subsubsection{The Role of Point Cloud Preprocessing}
As noted in the literature \cite{icra01}, there is a density difference between LiDAR and camera features; without any preprocessing of the point cloud, "only 5\% of camera features will be matched to a LiDAR point while all others will be dropped." By generating points in empty voxels, a denser point cloud can be achieved for intensive image feature sampling, alleviating this disparity. Farthest point sampling is only used in voxels with numerous LiDAR points, preserving the cloud's geometric features to some extent, reducing noise, and decreasing computational load during feature fusion. Omitting any of these methods would impair model accuracy, see \cref{table:my_label02}.
\subsubsection{Why Coordinate Concatenation}
Although 3D-to-2D projection is more stable than depth estimation given completely accurate intrinsic and extrinsic parameters, previous studies have indicated that inaccurate extrinsic calibration can impair feature fusion effectiveness~\cite{ral01}. We use deformable attention~\cite{iclr01} to mitigate calibration errors. Notably, if LiDAR voxel features are used directly as querys, our method resembles Spatial Cross Attention (SCA)~\cite{eccv01} in form. However, during feature fusion, we found that concatenating the coordinates of 3D reference points with LiDAR voxel features to form queries yields better results than using LiDAR voxel features alone, see \cref{table:my_label02}. This might be because the coordinate information of each point can provide distinct bias predictions for different reference points, thereby extracting the most effective camera features and point-by-point alleviating issues from imprecise calibration.
\subsubsection{Generalizability of Active Training}
Our proposed active training method is a very simple hard example mining technique. It's important to emphasize that this technique is designed to enhance the model's learning from challenging scenes, rather than biasing towards minority classes like loss functions such as focal loss~\cite{iccv05} do. Our training method is independent of loss function design and applicable across all models. This means that by using our training method in conjunction with loss functions biased towards minority classes, we can enable the model to learn from difficult regions within challenging samples from both macroscopic and microscopic perspectives, thereby further improving model performance. Experiments show that the active training method, with a resampling ratio of 70\%, simultaneously enhances the mIoU for both our OccFusion and M-CONet~\cite{iccv02}, demonstrating its high generalizability, see \cref{tab:my_label03}.

\begin{table}[tb]
    \centering
    \caption{Ablation study on the proposed methods. \emph{w/o} indicates without the corresponding method.}
    \begin{tabular}{c|C{2cm}}
     \toprule
        Methods & \makecell[c]{mIoU} \\
     \midrule
        w/o farthest point sampling & 21.8 \\
        w/o points generation & 21.2\\
        w/o coordinate concatenation & 22.0\\
        w/o active training method & 21.9\\
        w/o active coarse to fine & \textbf{22.5} \\
        OccFusion & 22.4\\
     \bottomrule
    \end{tabular}
    \label{table:my_label02}
\end{table}
\begin{table}[tb]
    \centering
    \caption{Ablation study on the resampling proportion in the resampling stage of the active training method.}
    \begin{tabular}{C{4cm}|C{3cm}|C{3cm}}
     \toprule
        Training resampling proportion (percent) & \makecell[c]{OccFusion mIoU} &  \makecell[c]{M-CONet mIoU}\\
     \midrule
        30 &  19.8 & 17.5\\
        50 & 22.0 & 20.3\\
        70 & \textbf{22.4} & \textbf{21.2}\\
     \bottomrule
    \end{tabular}
    \label{tab:my_label03}
\end{table}

\subsubsection{The Role of the Active Decoder}
In \cref{table:my_label02,tab:my_label04}, we observe that removing the entropy gate mechanism in the decoder results in a 0.1 improvement in the model's mIoU. However, we point out that for practical applications of occupancy prediction, reducing model complexity can enhance inference speed, necessitating a balance between accuracy and complexity. The entropy gate reduces the computational load by approximately 70\% during the feature refinement stage, achieving similar accuracy with significantly less complexity, making it a more efficient decoder. Moreover, like CONet\cite{iccv02}, our decoder does not impose any restrictions on the model's head and can be extended to any occupancy prediction model to enhance their performance.
\begin{table}[tb]
    \centering
    \caption{Ablation study on the proportion of voxels refined in the active coarse to fine pipeline}
    \begin{tabular}{C{4cm}|C{2cm}}
     \toprule
        Coarse to fine proportion (percent) & \makecell[c]{mIoU} \\
     \midrule
        10 &  17.9\\
        20 & 21.2\\
        30 & 22.4\\
        100 & \textbf{22.5}\\
     \bottomrule
    \end{tabular}
    \label{tab:my_label04}
\end{table}
\begin{table}[tb]
    \centering
    \caption{Experiment on the computational efficiency.  \emph{GPU Mem.} represents the GPU memory consumption at training phase. $\downarrow$: the lower, the better. $\uparrow$: the higher, the better.}
    \begin{tabular}{c|c|c|c}
     \toprule
        Methods  & GPU Mem.$(\downarrow)$ & GFLOPs $(\downarrow)$ & \makecell[c]{mIoU} $(\uparrow)$\\
     \midrule
         M-CONet & 24.0 GB & 3066 & 20.1 \\
         C-CONet & 22.0 GB & 2371 & 12.8 \\ 

        OccFusion  & \textbf{17.0} GB & \textbf{1566} & \textbf{22.4}\\
         
     \bottomrule
    \end{tabular}
    \label{tab:my_label05}
\end{table}
\subsubsection{Computational Performance of Our Model}
In \cref{tab:my_label05}, we showcase the computational performance of our OccFusion. Benefiting from our model's simpler mechanisms, it not only significantly enhances mIoU but also reduces the GFLOPs to half that of M-CONet. Notably, our model's GFLOPs are 34\% lower than those of C-CONet (note that C-CONet is a uni-modal method), while nearly doubling the mIoU. During the training phase, our model also uses fewer computational resources. This highlights our model's efficiency in maintaining low computational demand while ensuring reliability as a multi-modal model, demonstrating its exceptional performance.
\section{Conclusion}
In this paper, we introduce OccFusion, a depth estimation-free multi-modal fusion method that provides improved robustness over traditional methods. It incorporates a transferable active training method and an active occupancy decoder. Experiments on OpenOccupancy and Occ3D benchmarks confirm our method's superiority over current state-of-the-art models.Ablation studies further illustrates the effectiveness of our proposed components.

%
%
\bibliographystyle{splncs04}
\bibliography{main}

\begin{thebibliography}{10}
\providecommand{\url}[1]{\texttt{#1}}
\providecommand{\urlprefix}{URL }
\providecommand{\doi}[1]{https://doi.org/#1}

\bibitem{iccv01}
Behley, J., Garbade, M., Milioto, A., Quenzel, J., Behnke, S., Stachniss, C., Gall, J.: Semantickitti: A dataset for semantic scene understanding of lidar sequences. In: ICCV. pp. 9297--9307 (2019). \doi{10.1109/ICCV.2019.00939}

\bibitem{cvpr17}
Berman, M., Triki, A.R., Blaschko, M.B.: The lovasz-softmax loss: A tractable surrogate for the optimization of the intersection-over-union measure in neural networks. In: CVPR. pp. 4413--4421 (2018). \doi{10.1109/CVPR.2018.00464}

\bibitem{cvpr11}
Caesar, H., Bankiti, V., Lang, A.H., Vora, S., Liong, V.E., Xu, Q., Krishnan, A., Pan, Y., Baldan, G., Beijbom, O.: nuscenes: A multimodal dataset for autonomous driving. In: CVPR. pp. 11618--11628 (2020). \doi{10.1109/CVPR42600.2020.01164}

\bibitem{cvpr08}
Cao, A., de~Charette, R.: Monoscene: Monocular 3d semantic scene completion. In: CVPR. pp. 3981--3991 (2022). \doi{10.1109/CVPR52688.2022.00396}

\bibitem{cvpr03}
Chang, M.F., Lambert, J., Sangkloy, P., Singh, J., Bak, S., Hartnett, A., Wang, D., Carr, P., Lucey, S., Ramanan, D., Hays, J.: Argoverse: 3d tracking and forecasting with rich maps. In: CVPR. pp. 8748--8757 (2019). \doi{10.1109/CVPR.2019.00895}

\bibitem{cvpr18}
Chen, X., Lin, K., Qian, C., Zeng, G., Li, H.: 3d sketch-aware semantic scene completion via semi-supervised structure prior. In: CVPR. pp. 4192--4201 (2020). \doi{10.1109/CVPR42600.2020.00425}

\bibitem{cvpr14}
Deng, J., Dong, W., Socher, R., Li, L.J., Li, K., Fei-Fei, L.: Imagenet: A large-scale hierarchical image database. In: CVPR. pp. 248--255 (2009). \doi{10.1109/CVPR.2009.5206848}

\bibitem{tpami01}
Felzenszwalb, P.F., Girshick, R.B., McAllester, D.A., Ramanan, D.: Object detection with discriminatively trained part-based models. {IEEE} Trans. Pattern Anal. Mach. Intell.  \textbf{32}(9),  1627--1645 (2010). \doi{10.1109/TPAMI.2009.167}

\bibitem{cvpr01}
Firman, M., Aodha, O.M., Julier, S., Brostow, G.J.: Structured prediction of unobserved voxels from a single depth image. In: CVPR. pp. 5431--5440 (2016). \doi{10.1109/CVPR.2016.586}

\bibitem{icml01}
Gal, Y., Islam, R., Ghahramani, Z.: Deep bayesian active learning with image data. In: ICML. pp. 1183--1192 (2017)

\bibitem{cvpr13}
He, K., Zhang, X., Ren, S., Sun, J.: Deep residual learning for image recognition. In: CVPR. pp. 770--778 (2016). \doi{10.1109/CVPR.2016.90}

\bibitem{arxiv07}
Houlsby, N., Huszar, F., Ghahramani, Z., Lengyel, M.: Bayesian active learning for classification and preference learning. aeXiv preprint arXiv:1112.5745  (2011)

\bibitem{3dv01}
Hua, B.S., Pham, Q.H., Nguyen, D.T., Tran, M.K., Yu, L.F., Yeung, S.K.: Scenenn: A scene meshes dataset with annotations. In: 3DV. pp. 92--101 (2016). \doi{10.1109/3DV.2016.18}

\bibitem{arxiv12}
Huang, J., Huang, G.: Bevdet4d: Exploit temporal cues in multi-camera 3d object detection. CoRR  \textbf{abs/2203.17054} (2022). \doi{10.48550/ARXIV.2203.17054}

\bibitem{arxiv06}
Huang, J., Huang, G., Zhu, Z., Du, D.: Bevdet: High-performance multi-camera 3d object detection in bird-eye-view. aeXiv preprint arXiv:2112.11790  (2021)

\bibitem{cvpr02}
Huang, Y., Zheng, W., Zhang, Y., Zhou, J., Lu, J.: Tri-perspective view for vision-based 3d semantic occupancy prediction. In: CVPR. pp. 9223--9232 (2023). \doi{10.1109/CVPR52729.2023.00890}

\bibitem{iv01}
Kim, J., Choi, J., Kim, Y., Koh, J., Chung, C.C., Choi, J.W.: Robust camera lidar sensor fusion via deep gated information fusion network. In: IV. pp. 1620--1625 (2018). \doi{10.1109/IVS.2018.8500711}

\bibitem{iclr02}
Kingma, D.P., Ba, J.: Adam: A method for stochastic optimization. In: ICLR (2015)

\bibitem{neurips03}
Kirsch, A., van Amersfoort, J., Gal, Y.: Batchbald: Efficient and diverse batch acquisition for deep bayesian active learning. In: NeurIPS. pp. 7024--7035 (2019)

\bibitem{cvpr19}
Li, J., Han, K., Wang, P., Liu, Y., Yuan, X.: Anisotropic convolutional networks for 3d semantic scene completion. In: CVPR. pp. 3348--3356 (2020). \doi{10.1109/CVPR42600.2020.00341}

\bibitem{cvpr07}
Li, Y., Yu, Z., Choy, C.B., Xiao, C., Álvarez, J.M., Fidler, S., Feng, C., Anandkumar, A.: Voxformer: Sparse voxel transformer for camera-based 3d semantic scene completion. In: CVPR. pp. 9087--9098 (2023). \doi{10.1109/CVPR52729.2023.00877}

\bibitem{cvpr16}
Li, Y., Yu, A.W., Meng, T., Caine, B., Ngiam, J., Peng, D., Shen, J., Lu, Y., Zhou, D., Le, Q.V., Yuille, A.L., Ta, M.: Deepfusion: Lidar-camera deep fusion for multi-modal 3d object detection. In: CVPR. pp. 17161--17170 (2022). \doi{10.1109/CVPR52688.2022.01667}

\bibitem{aaai03}
Li, Y., Bao, H., Ge, Z., Yang, J., Sun, J., Li, Z.: Bevstereo: Enhancing depth estimation in multi-view 3d object detection with temporal stereo. In: Williams, B., Chen, Y., Neville, J. (eds.) AAAI. pp. 1486--1494. {AAAI} Press (2023). \doi{10.1609/AAAI.V37I2.25234}

\bibitem{eccv01}
Li, Z., Wang, W., Li, H., Xie, E., Sima, C., Lu, T., Yu, Q., Dai, J.: Bevformer: Learning bird's-eye-view representation from multi-camera images via spatiotemporal transformers. In: ECCV. pp. 1--18 (2022). \doi{10.1007/978-3-031-20077-9_1}

\bibitem{arxiv10}
Li, Z., Yu, Z., Austin, D., Fang, M., Lan, S., Kautz, J., {\'{A}}lvarez, J.M.: {FB-OCC:} 3d occupancy prediction based on forward-backward view transformation. CoRR  \textbf{abs/2307.01492} (2023). \doi{10.48550/ARXIV.2307.01492}

\bibitem{arxiv05}
Li, Z., Yu, Z., Austin, D., Fang, M., Lan, S., Kautz, J., Álvarez, J.M.: Fb-occ: 3d occupancy prediction based on forward-backward view transformation. aeXiv preprint arXIv:2307.01492  (2023). \doi{10.48550/arXiv.2307.01492}

\bibitem{cvpr21}
Lin, T., Doll{\'{a}}r, P., Girshick, R.B., He, K., Hariharan, B., Belongie, S.J.: Feature pyramid networks for object detection. In: 2017 {IEEE} Conference on Computer Vision and Pattern Recognition, {CVPR} 2017, Honolulu, HI, USA, July 21-26, 2017. pp. 936--944. {IEEE} Computer Society (2017). \doi{10.1109/CVPR.2017.106}

\bibitem{iccv05}
Lin, T., Goyal, P., Girshick, R.B., He, K., Doll{\'{a}}r, P.: Focal loss for dense object detection. In: ICCV. pp. 2999--3007. {IEEE} Computer Society (2017). \doi{10.1109/ICCV.2017.324}

\bibitem{ACM01}
Liu, P., Wang, L., Ranjan, R., He, G., Zhao, L.: A survey on active deep learning: From model driven to data driven. ACM Comput. Surv.  \textbf{54}(10s),  221:1--221:34 (2022). \doi{10.1145/3510414}

\bibitem{eccv02}
Liu, Y., Wang, T., Zhang, X., Sun, J.: Petr: Position embedding transformation for multi-view 3d object detection. In: ECCV. pp. 531--548 (2022). \doi{10.1007/978-3-031-19812-0_31}

\bibitem{icra01}
Liu, Z., Tang, H., Amini, A., Yang, X., Mao, H., Rus, D.L., Han, S.: Bevfusion: Multi-task multi-sensor fusion with unified bird's-eye view representation. In: {IEEE} International Conference on Robotics and Automation, {ICRA} 2023, London, UK, May 29 - June 2, 2023. pp. 2774--2781 (2023). \doi{10.1109/ICRA48891.2023.10160968}

\bibitem{arxiv09}
Lu, Y., Zhu, X., Wang, T., Ma, Y.: Octreeocc: Efficient and multi-granularity occupancy prediction using octree queries. CoRR  \textbf{abs/2312.03774} (2023). \doi{10.48550/ARXIV.2312.03774}

\bibitem{arxiv04}
Miao, R., Liu, W., Chen, M., Gong, Z., Xu, W., Hu, C., Zhou, S.: Occdepth: A depth-aware method for 3d semantic scene completion. arXiv preprint arXiv:2302.13540  (2023). \doi{10.48550/arXiv.2302.13540}

\bibitem{arxiv03}
Min, C., Xiao, L., Zhao, D., Nie, Y., Dai, B.: Uniscene: Multi-camera unified pre-training via 3d scene reconstruction. arXiv preprint arXiv:2305.18829  (2023). \doi{10.48550/arXiv.2305.18829}

\bibitem{arxiv08}
Ming, Z., Berrio, J.S., Shan, M., Worrall, S.: Occfusion: {A} straightforward and effective multi-sensor fusion framework for 3d occupancy prediction. CoRR  \textbf{abs/2403.01644} (2024). \doi{10.48550/ARXIV.2403.01644}

\bibitem{ral01}
Pan, J., Wang, Z., Wang, L.: Co-occ: Coupling explicit feature fusion with volume rendering regularization for multi-modal 3d semantic occupancy prediction. {IEEE} Robotics Autom. Lett.  \textbf{9}(6),  5687--5694 (2024). \doi{10.1109/LRA.2024.3396092}

\bibitem{arxiv13}
Pan, M., Liu, J., Zhang, R., Huang, P., Li, X., Liu, L., Zhang, S.: Renderocc: Vision-centric 3d occupancy prediction with 2d rendering supervision. CoRR  \textbf{abs/2309.09502} (2023). \doi{10.48550/ARXIV.2309.09502}

\bibitem{eccv03}
Philion, J., Fidler, S.: Lift, splat, shoot: Encoding images from arbitrary camera rigs by implicitly unprojecting to 3d. In: ECCV. pp. 194--210 (2020). \doi{10.1007/978-3-030-58568-6\_12}

\bibitem{cvpr05}
Qi, C.R., Zhou, Y., Najibi, M., Sun, P., Vo, K., Deng, B., Anguelov, D.: Offboard 3d object detection from point cloud sequences. In: CVPR. pp. 6134--6144 (2021). \doi{10.1109/CVPR46437.2021.00607}

\bibitem{neurips01}
Qi, C.R., Yi, L., Su, H., Guibas, L.J.: Pointnet++: Deep hierarchical feature learning on point sets in a metric space. In: NeurIPS. pp. 5099--5108 (2017)

\bibitem{cvpr06}
Reading, C., Harakeh, A., Chae, J., Waslander, S.L.: Categorical depth distribution network for monocular 3d object detection. In: CVPR. pp. 8555--8564 (2021). \doi{10.1109/CVPR46437.2021.00845}

\bibitem{3dv02}
Rold{\~{a}}o, L., de~Charette, R., Verroust{-}Blondet, A.: Lmscnet: Lightweight multiscale 3d semantic completion. In: 3DV. pp. 111--119 (2020). \doi{10.1109/3DV50981.2020.00021}

\bibitem{cvpr20}
Shrivastava, A., Gupta, A., Girshick, R.B.: Training region-based object detectors with online hard example mining. In: 2016 {IEEE} Conference on Computer Vision and Pattern Recognition, {CVPR} 2016, Las Vegas, NV, USA, June 27-30, 2016. pp. 761--769 (2016). \doi{10.1109/CVPR.2016.89}

\bibitem{phd01}
Sung, K.K.: Learning and example selection for object and pattern detection. Ph.D. thesis, Massachusetts Institute of Technology, Cambridge, MA, {USA} (1995), \url{https://hdl.handle.net/1721.1/9836}

\bibitem{neurips04}
Tian, X., Jiang, T., Yun, L., Mao, Y., Yang, H., Wang, Y., Wang, Y., Zhao, H.: Occ3d: {A} large-scale 3d occupancy prediction benchmark for autonomous driving. In: NeurIPS. pp. 64318--64330 (2023)

\bibitem{arxiv02}
Vobecky, A., Siméoni, O., Hurych, D., Gidaris, S., Bursuc, A., Pérez, P., Sivic, J.: Pop-3d: Open-vocabulary 3d occupancy prediction from images. arXiv preprint arXiv:2401.09413  (2024). \doi{10.48550/arXiv.2401.09413}

\bibitem{cvpr15}
Vora, S., Lang, A.H., Helou, B., Beijbom, O.: Pointpainting: Sequential fusion for 3d object detection. In: CVPR. pp. 4603--4611 (2020). \doi{10.1109/CVPR42600.2020.00466}

\bibitem{iccv02}
Wang, X., Zhu, Z., Xu, W., Zhang, Y., Wei, Y., Chi, X., Ye, Y., Du, D., Lu, J., Wang, X.: Openoccupancy: A large scale benchmark for surrounding semantic occupancy perception. In: ICCV. pp. 17804--17813 (2023). \doi{10.1109/ICCV51070.2023.01636}

\bibitem{cvpr04}
Wang, Y., Chao, W.L., Garg, D., Hariharan, B., Campbell, M., Weinberger, K.Q.: Pseudo-lidar from visual depth estimation: Bridging the gap in 3d object detection for autonomous driving. In: CVPR. pp. 8445--8453 (2019). \doi{10.1109/CVPR.2019.00864}

\bibitem{arxiv11}
Wang, Y., Chen, Y., Liao, X., Fan, L., Zhang, Z.: Panoocc: Unified occupancy representation for camera-based 3d panoptic segmentation. CoRR  \textbf{abs/2306.10013} (2023). \doi{10.48550/ARXIV.2306.10013}

\bibitem{iccv03}
Wei, Y., Zhao, L., Zheng, W., Zhu, Z., Zhou, J., Lu, J.: Surroundocc: Multi-camera 3d occupancy prediction for autonomous driving. In: ICCV. pp. 21672--21683 (2023). \doi{10.1109/ICCV51070.2023.01986}

\bibitem{aaai02}
Yan, X., Gao, J., Li, J., Zhang, R., Li, Z., Huang, R., Cui, S.: Sparse single sweep lidar point cloud segmentation via learning contextual shape priors from scene completion. In: AAAI. pp. 3101--3109 (2021). \doi{10.1609/AAAI.V35I4.16419}

\bibitem{sensor01}
Yan, Y., Mao, Y., Li, B.: Second: Sparsely embedded convolutional detection. Sensors  \textbf{18}(10) (2018). \doi{10.3390/S18103337}

\bibitem{cvpr09}
Yang, C., Chen, Y., Tian, H., Tao, C., Zhu, X., Zhang, Z., Huang, G., Li, H., Qiao, Y., Lu, L., Zhou, J., Dai, J.: Bevformer v2: Adapting modern image backbones to bird's-eye-view recognition via perspective supervision. In: CVPR. pp. 17830--17839 (2023). \doi{10.1109/CVPR52729.2023.01710}

\bibitem{neurips02}
Yin, T., Zhou, X., Kr{\"{a}}henb{\"{u}}hl, P.: Multimodal virtual point 3d detection. In: NeurIPS. pp. 16494--16507 (2021)

\bibitem{arxiv01}
Zhang, C., Yan, J., Wei, Y., Li, J., Liu, L., Tang, Y., Duan, Y., Lu, J.: Occnerf: Self-supervised multi-camera occupancy prediction with neural radiance fields. arXiv preprint arXiv:2312.09243  (2023). \doi{10.48550/arXiv.2312.09243}

\bibitem{aaai01}
Zhang, Y., Zheng, W., Zhu, Z., Huang, G., Lu, J., Zhou, J.: A simple baseline for multi-camera 3d object detection. In: AAAI. pp. 3507--3515 (2023). \doi{10.1609/aaai.v37i3.25460}

\bibitem{iccv04}
Zhang, Y., Zhu, Z., Du, D.: Occformer: Dual-path transformer for vision-based 3d semantic occupancy prediction. In: ICCV. pp. 9399--9409 (2023). \doi{10.1109/ICCV51070.2023.00865}

\bibitem{cvpr10}
Zhou, B., Krähenbühl, P.: Cross-view transformers for real-time map-view semantic segmentation. In: CVPR. pp. 13750--13759 (2022). \doi{10.1109/CVPR52688.2022.01339}

\bibitem{cvpr12}
Zhou, Y., Tuzel, O.: Voxelnet: End-to-end learning for point cloud based 3d object detection. In: CVPR. pp. 4490--4499 (2018). \doi{10.1109/CVPR.2018.00472}

\bibitem{iclr01}
Zhu, X., Su, W., Lu, L., Li, B., Wang, X., Dai, J.: Deformable detr: Deformable transformers for end-to-end object detection. In: ICLR (2021)

\end{thebibliography}
\end{document}